\documentclass[conference]{IEEEtran}
\IEEEoverridecommandlockouts
\usepackage{cite}
\usepackage{amsmath,amssymb,amsfonts}
\usepackage{graphicx}
\usepackage{textcomp}
\usepackage[colorlinks,linkcolor=red,anchorcolor=blue,citecolor=blue]{hyperref}
\usepackage{url}            % simple URL typesetting
\usepackage{booktabs}       % professional-quality tables
\usepackage{nicefrac}       % compact symbols for 1/2, etc.
\usepackage{microtype}      % microtypography
\usepackage[table,xcdraw]{xcolor}         % colors
\usepackage{multirow}
\usepackage{makecell}
\usepackage{caption}
\usepackage{pifont}
\usepackage{threeparttable}
\usepackage{subfigure}
\usepackage{subcaption}
\usepackage[T1]{fontenc}
\usepackage{courier}
\usepackage{bm}
\usepackage{algorithm}
\usepackage{algpseudocodex}
\usepackage[switch,columnwise]{lineno}
\usepackage{bm}

\DeclareMathAlphabet\mathbfcal{OMS}{cmsy}{b}{n}
\newcommand{\ten}[1]{\mathbfcal{#1}} %mathcal
\newcommand{\mat}[1]{\mathbf{#1}}

\begin{document}
% \linenumbers

\title{%
Tensor-Compressed and Fully-Quantized Training of Neural PDE Solvers
% on Edge Devices
%   \thanks{This work is co-funded by Intel Strategic Research Sectors (SRS) - Systems
% Integration SRS \& Devices SRS. (Corresponding author: Zheng Zhang.)}
}

\author{\IEEEauthorblockN{Jinming Lu}
  \IEEEauthorblockA{
    \textit{UC Santa Barbara }\\
    Santa Barbara, USA \\
    jinminglu@ucsb.edu}
  \and

  \IEEEauthorblockN{Jiayi Tian}
  \IEEEauthorblockA{
    \textit{UC Santa Barbara} \\
    Santa Barbara, USA \\
    jiayi\_tian@ucsb.edu}
  \and

  \IEEEauthorblockN{Yequan Zhao}
  \IEEEauthorblockA{
    \textit{UC Santa Barbara} \\
    Santa Barbara, USA \\
    yequan\_zhao@ucsb.edu}
  \and

    \IEEEauthorblockN{Hai Li}
  \IEEEauthorblockA{
    \textit{Intel Corporation} \\
    Portland, USA \\
    hai.li@intel.com}
  \and

  \IEEEauthorblockN{Zheng Zhang}
  \IEEEauthorblockA{
    \textit{UC Santa Barbara} \\
    Santa Barbara, USA \\
    zhengzhang@ece.ucsb.edu}
}

\maketitle

\begin{abstract}
  Physics-Informed Neural Networks (PINNs) have emerged as a promising paradigm for solving partial differential equations (PDEs) by embedding physical laws into neural network training objectives. However, their deployment on resource-constrained platforms is hindered by substantial computational and memory overhead, primarily stemming from  higher-order automatic differentiation, intensive tensor operations, and reliance on full-precision arithmetic.
  To address these challenges, we present a framework that enables scalable and energy-efficient PINN training on edge devices.
  This framework integrates fully quantized training, Stein's estimator (SE)-based residual loss computation, and tensor-train (TT) decomposition for weight compression.
  It contributes three key innovations:
  (1) a mixed-precision training method that use a square-block MX (SMX) format to eliminate data duplication during backpropagation;
  (2) a difference-based quantization scheme for the Stein's estimator that mitigates underflow;
  and (3) a partial-reconstruction scheme (PRS) for TT-Layers that reduces quantization-error accumulation.
  We further design PINTA, a precision-scalable hardware accelerator, to fully exploit the performance of the framework.
  Experiments on the 2-D Poisson, 20-D Hamilton--Jacobi--Bellman (HJB), and 100-D Heat equations demonstrate that the proposed framework achieves accuracy comparable to or better than full-precision, uncompressed baselines while delivering $5.5\times$ to $83.5\times$ speedups and $159.6\times$ to $2324.1\times$ energy savings. This work enables real-time PDE solving on edge devices and paves the way for energy-efficient scientific computing at scale.
  \end{abstract}


\section{Introduction}
Physics-Informed Neural Networks (PINNs) \cite{raissi2019physics} have emerged as a promising approach to solving partial differential equations (PDE) by embedding physical laws directly into the training objective of neural networks.
They have been successfully applied across diverse domains, including inverse-scattering problems in nano-optics \cite{chen2020physics, chen2022physics}, thermomechanical modeling \cite{nguyen2022physics}, and control of dynamical systems in robotics \cite{antonelo2024physics, velioglu2025physics}. Unlike traditional PDE solvers, PINNs leverage data-driven learning and automatic differentiation (AD) to enforce PDE constraints, yielding mesh-free solutions that generalize across problem settings.

Despite these advantages, training PINNs is computationally and memory intensive. This is primarily due to three factors: (1) reliance on higher-order AD to compute PDE residuals; (2) the large model sizes required to capture complex physical behavior; and (3) pervasive use of high-precision floating-point arithmetic during training because of the higher sensitivity of PINN with respect to quantization errors.
For example, second-order PDEs require computing and storing numerous Jacobian and Hessian terms at every collocation point \cite{he2023learning, hu2024hutchinson}, often consuming $10\times$ to $100\times$ more memory than standard neural-network training. These costs are a major barrier to deploying PINNs on resource-contained edge platforms (e.g., autonomous robotics and embedded scientific instrumentation), where memory, latency, and power budgets are tightly limited.

To mitigate the computational and memory burdens, prior work has proposed algorithmic and hardware-focused solutions.
For instance, methods such as \cite{liu2022tt, vemuri2025functional} leverage tensor decomposition to compress network weights and reducing model size. However, these methods remain software-only and still rely on full-precision AD, providing limited end-to-end acceleration.
Other efforts, such as \cite{zhao2025scalable, zhaoreal} propose a backpropagation-free training framework that employs zeroth-order optimization to eliminate AD and gradient backpropagation.
These methods are implemented on photonic hardware accelerators, achieving notable gains in efficiency. Nevertheless, their reliance on specialized photonic ASICs restricts portability and broader adoption on conventional digital platforms.

% Furthermore, existing hardware accelerators and training frameworks are not well-optimized for the unique characteristics of PINNs, leading to suboptimal performance.

% Moreover, current hardware accelerators and training frameworks are poorly suited to the unique computational patterns of PINNs—particularly their frequent use of high-order derivatives and large intermediate tensor allocations—leading to suboptimal runtime and energy performance.

Motivated by these limitations, we propose a holistic framework that enables scalable and energy-efficient PINN training on edge devices. Our key contributions are as follows.
\begin{itemize}
    \item We propose an efficient on-device PINN training framework by integrating three core techniques: fully quantized training, Stein's estimator (SE)-based residual loss computation, and tensor-train (TT) decomposition. The framework features a novel mixed-precision strategy that leverages Square-block MX-INT (SMX) formats to avoid redundant data duplication while preserving representational fidelity.
    \item We introduce two accuracy-preserving techniques: (i) a difference-based quantization method (DiffQuant) that decouples quantization noise from perturbations within Stein's estimator computation; and (ii) a partial-reconstruction scheme (PRS) for TT-Layers that mitigates quantization error accumulation across tensor contractions.
    \item We design and evaluate a precision-scalable hardware accelerator optimized for the proposed training pipeline. Implemented on 7-nm technology, the design provides up to $83.5\times$ speedup and $2324.1\times$ energy reduction compared to  an AD-based baseline, and $18.3\times$ speedup and $16.0\times$ energy reduction compared to an SE-based baseline.
\end{itemize}
This work brings PINNs closer to practical deployment in constrained environments, opening the door to real-time PDE solving on edge devices and energy-efficient scientific modeling at scale.

% Together, these innovations form an efficient and scalable solution for PINN training, enabling broader adoption of physics-informed learning in real-world applications.
% Together, these innovations reduce both training time and energy consumption while maintaining solution accuracy. Our hardware implementation demonstrates providing $5.5\times\sim 83.5\times$ speedup and $159.6\times\sim 2324.1\times$ energy reduction compared to a baseline full-precision AD-based PINN training pipeline.
% This work brings PINNs closer to practical deployment in constrained environments, opening the door to real-time PDE solving on edge devices and energy-efficient scientific modeling at scale.

\section{Background}

\subsection{Physics-Informed Neural Networks}

\begin{figure}[tb]
  \centering
  \includegraphics[width=0.5\textwidth]{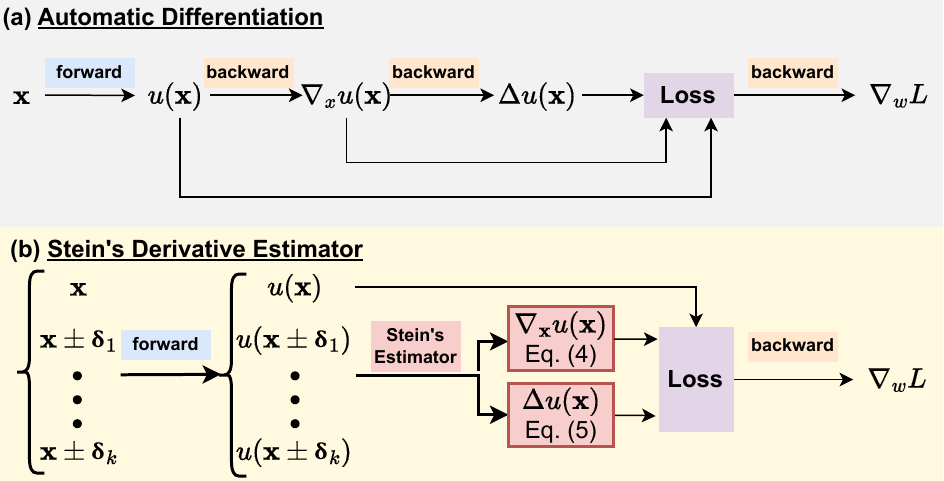}
  \caption{Training flow of PINN with (a) Automatic Differentiation and (b) Stein's Estimator. }
  \label{fig:stein}
  \vspace{-10pt}
\end{figure}

Physics-Informed Neural Networks (PINNs) \cite{raissi2019physics} are a class of deep learning models that integrate physical knowledge into the architecture of neural networks to solve forward and inverse problems of PDEs. Formally, we consider a generic PDE defined over a $D$-dimensional domain $\Omega \subset \mathbb{R}^D$:
\begin{align}
    \mathcal{F}[u(\mathbf{x})] &= 0, \quad \mathbf{x} \in \Omega, \\
    \mathcal{B}[u(\mathbf{x})] &= 0, \quad \mathbf{x} \in \partial\Omega,
\end{align}
where \( \mathcal{F} \) denotes a differential operator and \( \mathcal{B} \) enforces boundary or initial conditions. A neural network \( u_{\mathbf{\theta}}(\mathbf{x}) \), parameterized by weights \( \mathbf{\theta} \), is trained to approximate the solution \( u(\mathbf{x}) \). The training loss typically consists of physics-driven and optionally data-driven components:
\begin{equation}
\begin{aligned}
\mathcal{L}(\theta) = &\, \frac{w_c}{N_c} \sum_{i=1}^{N_c} \left\| \mathcal{F}[u_\theta(\mathbf{x}_c^i)] \right\|^2
+ \frac{w_b}{N_b} \sum_{i=1}^{N_b} \left\| \mathcal{B}[u_\theta(\mathbf{x}_b^i)] \right\|^2 \\
&+ \frac{w_d}{N_d} \sum_{i=1}^{N_d} \left\| u_\theta(\mathbf{x}_d^i) - u(\mathbf{x}_d^i) \right\|^2,
\end{aligned}
\end{equation}
where $w_c, w_b, w_d$ are loss weights, and $N_c, N_b, N_d$ denote the number of data points respectively. The first two terms correspond to PDE residuals loss and boundary/initial condition loss, while the third is the regular data loss to fit the dataset.  PINNs typically employ automatic differentiation (AD) to compute derivatives  required for enforcing PDE constraints. However, as shown in Fig. \ref{fig:stein}(a), computing high-order derivatives via AD entails repeated backpropagation, making the process both memory- and compute-intensive, especially for high-dimensional or higher-order PDEs.

\subsubsection{Stein's Derivative Estimator for PDE Residuals}
To alleviate the computational burden of AD,
He et al. \cite{he2023learning} introduced Stein’s derivative estimator (SE), a sampling-based forward-mode technique for derivative approximation. As shown in Fig. \ref{fig:stein} (b), it enables derivative computation without explicit backpropagation.
% \zz{I think it's a biased estimator. Please double check} \jm{It's a unbiased estimator by introduce $\pm \delta$ perturbation}

For a differentiable function \( u: \mathbb{R}^d \rightarrow \mathbb{R} \), the first-order derivative can be estimated via:
\begin{equation}
  \begin{aligned}
  \nabla_\mat{x} u(\mat{x}) &= \mathbb{E}_{\bm{\delta} } \left[ \frac{\bm{\delta}}{2\sigma^2} \left( u(\mat{x} + \bm{\delta}) - u(\mat{x} - \bm{\delta}) \right) \right] \\
                & \approx \frac{1}{K} \sum_{i=1}^{K} \frac{\bm{\delta}_i}{2\sigma^2} \left( u(\mat{x} + \bm{\delta}_i) - u(\mat{x} - \bm{\delta}_i) \right),
  \end{aligned}
\end{equation}
% \zz{Need to specify the distribution of $\delta$. Use bold font to represent vectors (e.g. $\mat{x}$) and matrices (e.g., $\mat{A}$) to make the notations consistent in the whole paper.} \jm{Fixed}
where $\bm{\delta}$ is a random perturbation sampled from $\mathcal{N}(0, \sigma^2 \mat{I})$.
Higher-order derivatives such as the Laplacian can also be approximated through Eq.~\eqref{eq:laplac}.
%using higher-moment extensions of Stein’s identity:
\begin{equation}\label{eq:laplac}
  \begin{aligned}
% H f(x) &= \mathbb{E}_{\delta \sim \mathcal{N}(0, \sigma^2 I)} \left[ \frac{\delta\delta^T - \sigma^2 I}{\sigma^4} f(x + \delta) \right], \\
\Delta u(\mat{x}) = \mathbb{E}_{\bm{\delta}} \left[ \frac{\|\bm{\delta}\|^2 - \sigma^2 D}{2 \sigma^4}   \left( u \left(\mat{x} + \bm{\delta} \right) + u \left(\mat{x} - \bm{\delta} \right) - 2 u(\mat{x}) \right) \right] \\
            \approx \frac{1}{K} \sum_{i=1}^{K} \frac{\|\bm{\delta}_i\|^2 - \sigma^2 D}{2 \sigma^4} \cdot  \left( u \left(\mat{x} + \bm{\delta}_i \right) + u \left(\mat{x} - \bm{\delta}_i \right) - 2 u(\mat{x}) \right).
\end{aligned}
\end{equation}
% These identities allow the estimation of differential operators directly from function evaluations at perturbed inputs, thus completely bypassing back-propagation for derivative computations. Incorporating Stein's estimator into PINN training significantly reduces the computational burden of derivative calculation while preserving accuracy, particularly in high-dimensional PDE problems.
% This approach enables scalable and hardware-efficient learning of physics-informed models.

\begin{figure}[tb]
  \centering
  \includegraphics[width=0.5\textwidth]{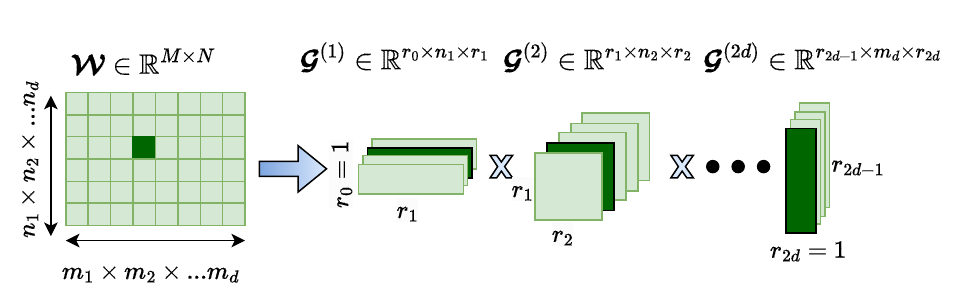}
  \caption{Illustration of Tensor-Train Decomposition.}
  \label{fig:tt}
    \vspace{-10pt}
\end{figure}

\begin{figure*}[!ptb]
  \centering
  \includegraphics[width=0.98\textwidth]{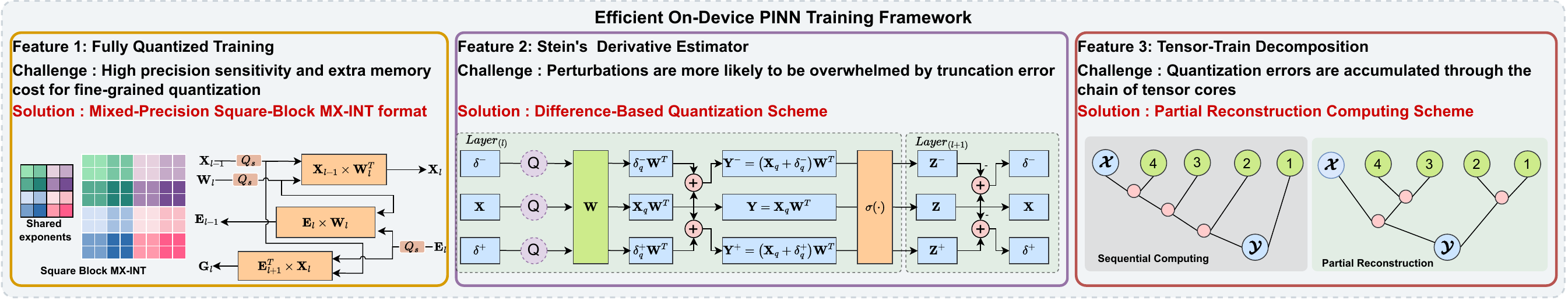}
  \caption{Overview of the proposed efficient on-device PINN training framework. }
  \label{fig:overview}
    \vspace{-10pt}
\end{figure*}

\subsection{Tensor-Train Decomposition}
Tensor-Train (TT) decomposition \cite{oseledets2011tensor,tian2025ultra} is a low-rank tensor factorization technique that efficiently represents high-dimensional tensors using a sequence of lower-dimensional core tensors.
Originally introduced to address the curse of dimensionality in numerical computations, TT decomposition has found wide applications in machine learning for model compression and efficient inference. In the context of neural networks, TT decomposition is typically applied to fully connected layers by reshaping weight matrices into high-order tensors, followed by a low-rank decomposition in a chain structure. This approach drastically reduces the number of parameters and enables memory- and compute-efficient implementation on hardware accelerators.
%Furthermore, TT layers retain the expressivity of dense layers when TT-ranks are chosen appropriately, making them a compelling choice for deploying large models under resource constraints.

Formally, consider a weight matrix $\mat{W} \in \mathbb{R}^{M \times N}$, which is first reshaped into a $2d$-dimensional tensor $\ten{W} \in \mathbb{R}^{m_1  \times \cdots \times m_d \times n_1  \times \cdots \times n_d}$, where $N = \prod_{i=1}^{d} n_i$ and $M = \prod_{i=1}^{d} m_i$.
As shown in Fig. \ref{fig:tt}, TT factorizes $\ten{W}$ into the product of $2d$ third-order core tensors $\{\ten{G}^{(k)}\}_{k=1}^{2d}$ such that:
\begin{equation}\label{eq:tt}
  \begin{aligned}
  &\ten{W}_{[ i_1, ... i_d, j_1, ... j_d]} = \\
  & \sum_{r_1...r_d} \ten{G}^{(1)}_{[r_0, i_1, r_1]}... \ten{G}^{(d)}_{[r_{d-1}, i_{d}, r_{d}]}
   \ten{G}^{(d+1)}_{[r_{d}, j_{1}, r_{d+1}]} ...\ten{G}^{(2d)}_{[r_{2d-1}, j_{d}, r_{2d}]}.
  \end{aligned}
\end{equation}
where $\ten{G}^{(k)} \in \mathbb{R}^{r_{k-1} \times m_k \times r_k}$ for $1 \leq k \leq d$, and $\ten{G}^{(k)} \in \mathbb{R}^{r_{k-1} \times n_{k-d} \times r_k}$ for $d < k \le 2d $.
   $\{r_k\}_{k=1}^{2d}$ are known as TT-ranks, which determine the compression rate and expressiveness of the decomposition.
By applying the TT decomposition, the standard linear layer $\mat{Y} = \mat{X} \mat{W}^T$ is replaced with a TT-Layer.
The TT-Layer can be expressed as:
\begin{equation}
  \ten{Y}_{[b,i_1\cdots i_d]} = \sum_{j_1 ... j_d}  \ten{G}^{(1)}{[i_1]} \ten{G}^{(2)}{[i_2]} ... \ten{G}^{(2d)}{[j_d]} \ten{X}_{[b,j_1 ... j_d]},
\end{equation}
where $b$ is the batch dimension, $\ten{G}^{(i)}[i_k] \in \mathbb{R}^{r_{i-1} \times r_i}$ is the $i_k$-th slice of the TT-core $\ten{G}^{(i)}$ by fixing its second index as $i_k$.

\section{Method}

\subsection{Overview}
In this work, we develop an efficient framework for training PINNs on resource-constrained devices.
To reduce the computational and memory overhead of PINN training, as illustrated in Fig.~\ref{fig:overview}, the framework integrates three key components: \textbf{fully quantized training} to reduce memory footprint and arithmetic cost,  \textbf{Stein's estimator} for low-cost derivative computation without backpropagation, and \textbf{TT decomposition} for compact weight representation.

However, naively combining these techniques into a cohesive, high-performance pipeline introduces several challenges.
\ding{202} Compared with models in vision and natural language processing, PINNs are particularly sensitive to quantization-induced errors, especially due to their use in scientific domains \cite{tu2023guaranteed,  Dool2023EfficientNP, hayford2024speeding}.
\ding{203} The small perturbations used by  SE-based training can be obscured by quantization noise, leading to inaccurate gradient estimates
\ding{204} Tensorization reduces parameter count but increases the number of tensor contractions, which can exacerbate the accumulation of quantization errors.

 We address these issues with {\bf three key innovations}: (1) a mixed-precision training strategy that uses Square-block MX-INT (SMX) format to balance accuracy and efficiency; (2) a difference-based quantization scheme (DiffQuant) to preserve the sensitivity of Stein's estimator under low-bit arithmetic; and (3) a partial-reconstruction scheme (PRS) for TT-Layers that minimizes quantization error accumulation while maintaining computational efficiency.
This design preserves the benefits of each component while mitigating adverse interactions among them.

\subsection{Fully Quantized Training for PINNs}

\begin{figure}[tb]
  \centering
  \includegraphics[width=0.5\textwidth]{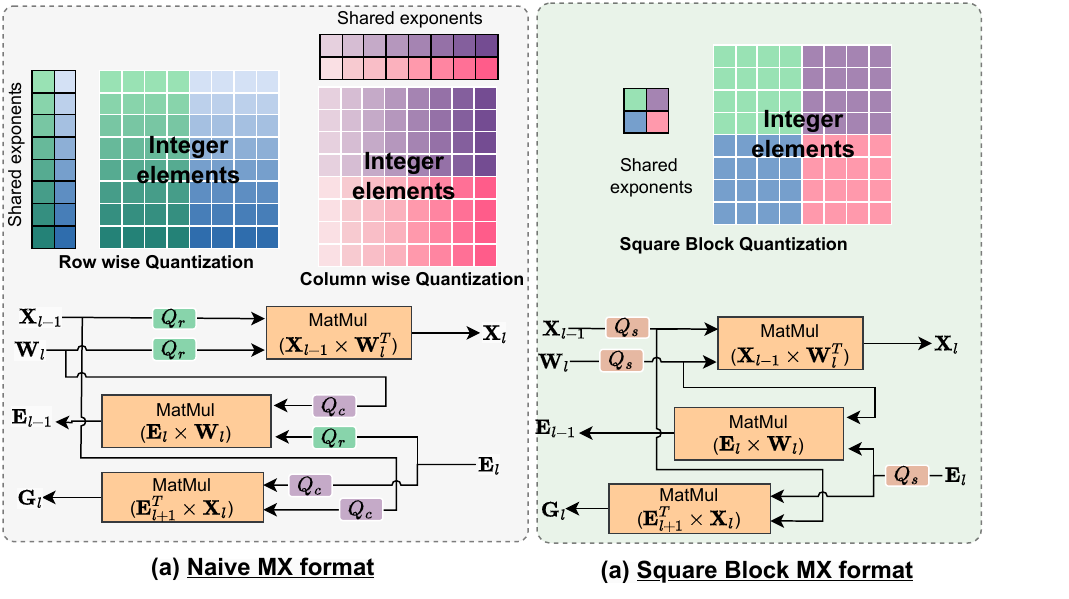}
  \caption{Computing flow of fully quantized training with (a) MX format and (b) Square-block MX format.   \vspace{-10pt}}
  \label{fig:mx}
\end{figure}

Microscaling (MX) data formats \cite{rouhani2023microscaling} are a class of quantization schemes that enable low-bit training and inference by assigning fine-grained per-block scaling factors. These formats have achieved excellent results across various vision and language models \cite{kim2024dacapo, zhang2025sageattention3}.
However, conventional MX formats are directional. As shown in Fig.~\ref{fig:mx}, this necessitates transposing and duplicating tensors during backpropagation, which increases both memory usage and runtime \cite{darvish2023shared}.

To address this limitation, we adopt the Square-block MX-INT (SMX) format, where each $4\times 4$ data block shares a common exponent \cite{cuyckens2025efficient}.
This bidirectional design eliminates the need for redundant copies and supports efficient forward and backward passes. The quantization process is defined as
\begin{equation}
    \begin{aligned}
    \text{shared\_exp}[i] &= \lfloor \log_2( \max \left( |\mat{X}[i]| \right)) - emax \rfloor, \\
    s[i] &= 2^{\text{shared\_exp}[i]}, \\
    \hat{\mat{X}}[i] &= s[i] \cdot \text{round}\left(\mat{X}[i]/s[i]\right), \\
    % \hat{\mat{X}}[i] &= s[i] \text{clamp}\left(\left[ \mat{X}[i]/s[i] \right], -2^{b-1}, 2^{b-1} - 1  \right) \\
    \end{aligned}
\end{equation}
where $\mat{X}[i]$ is $i$-th block of matrix $\mat{X}$, and $emax$ is the maximum exponent value.

Given the numerical sensitivity of PINN training, we further explore mixed precision across tensor types. Our empirical analysis shows that INT8 is sufficient for activations and weights, whereas INT12 is necessary for gradients.
% During training, operations such as matrix multiplications are performed on quantized MX representations, while non-dot product layers (e.g., activation functions and normalization) keep using the original precision.

\subsection{Difference-based Quantization for Stein's Estimator}

\begin{table}[t]
  \centering
  \begin{threeparttable}
    \renewcommand{\arraystretch}{1.2}
  \resizebox{0.85\columnwidth}{!}{%
  \begin{tabular}{c|c|c|c}
  \toprule
  \textbf{PDE Loss} & \textbf{Precision} & \textbf{MSE} & \textbf{$\ell_2$ Rel. Error} \\
  \hline
  AD            & FP32 & 1.15E-3   & 2.86E-3  \\ \hline
  \multirow{5}{*}{SE}         & FP32 & 1.59E-3  & 3.93E-3  \\
           & W8-A8-E8 & 1.26E-1 & 3.73E-1 \\
           & W8-A32-E32 & 1.67E-3 & 2.65E-3 \\
           & W32-A8-E32 & 1.15E-1 & 3.15E-1 \\
          %  & W32-A32-E8 & 0.00114 & 0.00283 \\
  \bottomrule
  \end{tabular}
  }
  \end{threeparttable}
  \caption{Comparison of PDE loss computation methods and numeric precision on PINN performance for Possion 2D.   \vspace{-10pt} }
  \label{tab:quant_ablation}
  \end{table}

% Quantizing Physics-Informed Neural Networks (PINNs) introduces several non-trivial challenges due to their reliance on precise gradient computations and fidelity to physical laws. First, low-precision arithmetic impairs the accurate evaluation of derivatives, which are critical for enforcing partial differential equation (PDE) constraints in the loss function. This loss of precision is especially detrimental when computing higher-order derivatives, often required in PINNs. Second, many physical systems governed by PDEs are sensitive to small perturbations, and quantization-induced noise can lead to instability or physically inconsistent solutions. Finally, empirical benchmarks and best practices for quantized PINNs remain underexplored, particularly under high-dimensional or nonlinear regimes.

The Stein’s estimator avoids explicit computation of higher-order derivatives of $u(\mat{x})$, simplifying the quantization workflow to align with standard training processes. However, our preliminary experiments reveal that the performance of a fully quantized model is unsatisfactory.
As shown in Table~\ref{tab:quant_ablation}, the quantized model exhibits significantly higher mean squared error (MSE) and $\ell_2$ relative error compared to its full-precision counterpart. Through ablation studies, we identify that the primary source of performance degradation stems from the loss of activation precision.

\begin{figure}[tb]
  \centering
    \includegraphics[width=0.5\textwidth]{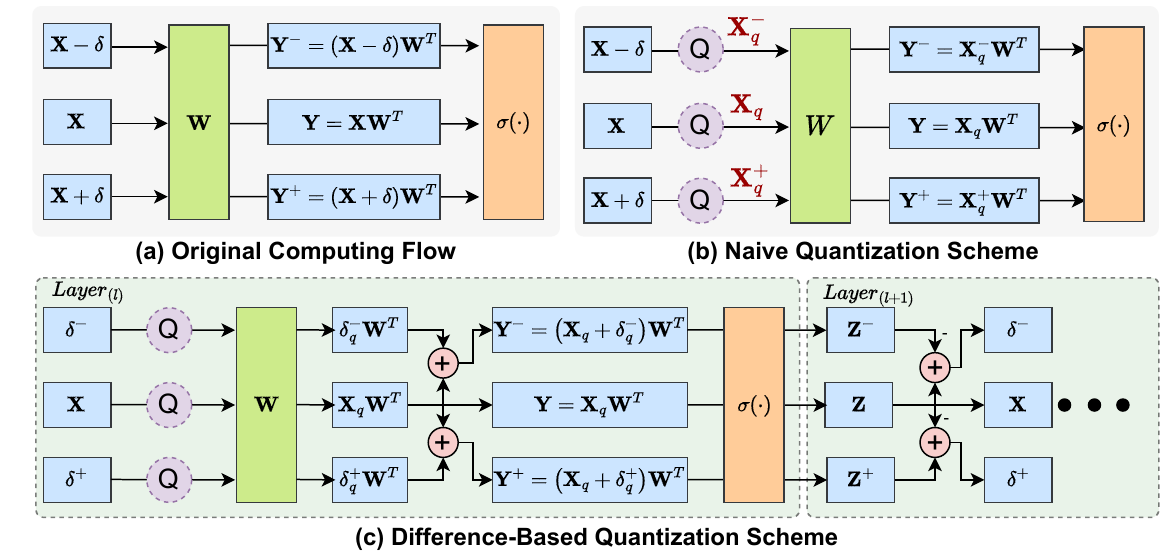}
  \caption{Computing flow of difference-based quantization scheme. }
  \label{fig:quant}
    \vspace{-10pt}
\end{figure}

This issue stems from the core mechanism of the Stein's estimator. As illustrated in Fig. \ref{fig:quant}(a), derivative estimates are computed from the difference between $u(\mat{X})$ and $u(\mat{X} \pm \boldsymbol{\delta})$, where $\boldsymbol{\delta} \sim \mathcal{N}(0, \sigma^2 \mat{I})$ is a small random perturbation (typically, $\sigma=0.01$).
In the quantized computation scheme shown in Fig.~\ref{fig:quant} (b), the injected noise is often smaller than the quantization step size and is consequently masked by quantization error.
As a result, the quantized perturbed input $\mat{X}_q^+ = Q(\mat{X} + \boldsymbol{\delta})$ frequently collapses to the same value as the quantized original input $\mat{X}_q = Q(\mat{X})$, thereby invalidating the gradient estimation via Stein’s method.

To formalize this ``quantization masking'' effect, we analyze the distinguishability of a scalar value $x_i$ from its perturbed version $x_i + \delta_i$.
Let \( Q(\cdot) \) denote a uniform quantizer with bit-width \( b \) and step size \( s = (x_{\max} - x_{\min})/({2^{b} - 1}) \). The quantized value is given by:
\begin{equation}
    Q(x_i) = s \cdot \text{round}\left( \frac{x_i}{s} \right).
\end{equation}
% can be expressed as:
% \begin{equation}
%     Q(x_i + \delta_i) - Q(x_i) = s \cdot \left( \text{round}\left( \frac{x_i + \delta_i}{s} \right) - \text{round}\left( \frac{x_i}{s} \right) \right).
% \end{equation}
The difference between the quantized values, $Q(x_i)$ and $Q(x_i + \delta_i)$, is non-zero only if the perturbation $\delta_i$ forces $x_i$ to cross the quantization threshold.
This event depends on both the magnitude of $\delta_i$ and the proximity of $x_i$ to a threshold.

To quantify this, we calculate the probability of such a ``quantization flip''.
Assuming the position of $x_i$ is uniformly distributed within it quantization bin, the distance $l$ from $x_i$ to the nearest quantization threshold follow a uniform distribution $l \sim \mathcal{U}[0, s/2]$. A flip occurs if $| \delta_i | > l$.
Given that $\delta_i\sim \mathcal{N}(0, \sigma^2$), the conditional probability of a flip is defined as:
\begin{equation}
\mathbb{P}(|\delta_i| > l) = 2 \cdot \Phi\left(-\frac{l}{\sigma}\right),
\end{equation}
where $\Phi(\cdot)$ is the cumulative distribution function (CDF) of the standard normal distribution.
By marginalizing over all possible values of $l$, the overall flip probability is:
 \begin{equation}
  \begin{aligned}
P_{\text{flip}} &= \mathbb{E}_{l \sim \mathcal{U}[0, s/2]}\left[ \mathbb{P}(|\delta| > l) \right] \\
&= \boxed{\frac{4}{s} \int_0^{s/2} \Phi\left( -\frac{l}{\sigma} \right) dl}.
\end{aligned}
\end{equation}
For 8-bit quantization of data normalized to the range of $[-1, 1]$, this yields a flip probability of approximately 15.5\%, implying that a significant portion of perturbed elements are indistinguishable after quantization. This quantization masking effect severely impairs the effectiveness of Stein’s estimators.

% which depends on both the alignment of $x_i$ within the quantization bin and the magnitude of $\delta_i$.

% To account for all possible alignments of $x_i$ within a quantization bin, we marginalize over $x_i \bmod s \sim \mathcal{U}[0, s]$. Define $l(x_i) = \min\left(|x_i - Q(x)|, s/2\right)$ as the distance from $x_i$ to the nearest quantization threshold. Since $x_i$ is uniformly distributed within the bin, $l \sim \mathcal{U}[0, s/2]$. A quantization flip occurs if $|\delta_i| > l$, and thus the conditional probability of a flip given $l$ is:
% \begin{equation}
% \mathbb{P}_{\delta_i}(|\delta_i| > l) = 2 \cdot \Phi\left(-\frac{l}{\sigma}\right),
% \end{equation}
% where $\Phi(\cdot)$ denotes the cumulative distribution function of the standard normal distribution. Therefore, the overall flip probability is given by:
% \begin{equation}
%   \begin{aligned}
% P_{\text{flip}} &= \mathbb{E}_{l \sim \mathcal{U}[0, s/2]}\left[ \mathbb{P}(|\delta| > l) \right] \\
% &= \boxed{\frac{4}{s} \int_0^{s/2} \Phi\left( -\frac{l}{\sigma} \right) dl}.
% \end{aligned}
% \end{equation}
% For 8-bit quantization under normalized data in the range of $[-1, 1]$, this yields a flip probability of approximately 15.5\%, implying that a significant portion of perturbed elements are indistinguishable after quantization. This quantization masking effect severely impairs the effectiveness of Stein’s estimators.

To resolve the quantization masking issue, we propose a difference-based quantization scheme, DiffQuant. Instead of quantizing the perturbed activation $\mat{X} + \boldsymbol{\delta}$ as a single term, our method quantizes the original activation $\mat{X}$ and the perturbation $\boldsymbol{\delta}$ separately.

As illustrated in Fig.~\ref{fig:quant}(c), the forward pass of a perturbed input through a linear layer can be decomposed:
\begin{equation}
\mat{Y}^+ = (\mat{X} + \boldsymbol{\delta})\mat{W}^T + \mat{b} = \mat{X}\mat{W}^T + \boldsymbol{\delta}\mat{W}^T + \mat{b}.
\end{equation}
This decomposition allows us to replace the naive quantization approach with our DiffQuant method (weight quantization is omitted for clarity):
\begin{equation}
\begin{aligned}
\mat{Y}^+ &= Q(\mat{X} + \boldsymbol{\delta})\mat{W}^T + \mat{b} & \text{(NaiveQuant)} \\
\mat{Y}^+ &= Q(\mat{X})\mat{W}^T + Q(\boldsymbol{\delta})\mat{W}^T + \mat{b} & \text{(DiffQuant)}
\end{aligned}
\end{equation}
This formulation decouples the quantization of the perturbation $\boldsymbol{\delta}$ from the base activation $\mat{X}$.
By doing so, it prevents the quantization error of $\mat{X}$ from overwhelming the small perturbation signal, thereby preserving the accuracy of gradient estimates from the Stein's method.

The perturbation must also be propagated through nonlinear activation functions, $\sigma(·)$. For the subsequent layer (l+1), the new perturbation is not $\sigma(Q(\boldsymbol{\delta_l}))$ but is instead recomputed as the difference between the activated outputs:
\begin{equation}
\begin{aligned}
\boldsymbol{\delta}^+_{l+1} &= \sigma(\mat{Y}^+) - \sigma(\mat{Y}), \\
\boldsymbol{\delta}^-_{l+1} &= \sigma(\mat{Y}) - \sigma(\mat{Y}^-).
\end{aligned}
\end{equation}
These newly computed perturbation terms, $\boldsymbol{\delta}^+_{l+1}$ and $\boldsymbol{\delta}^-_{l+1}$, are then used to estimate the derivatives for the next layer.

\begin{figure}[tb]
  \centering
  \includegraphics[width=\columnwidth]{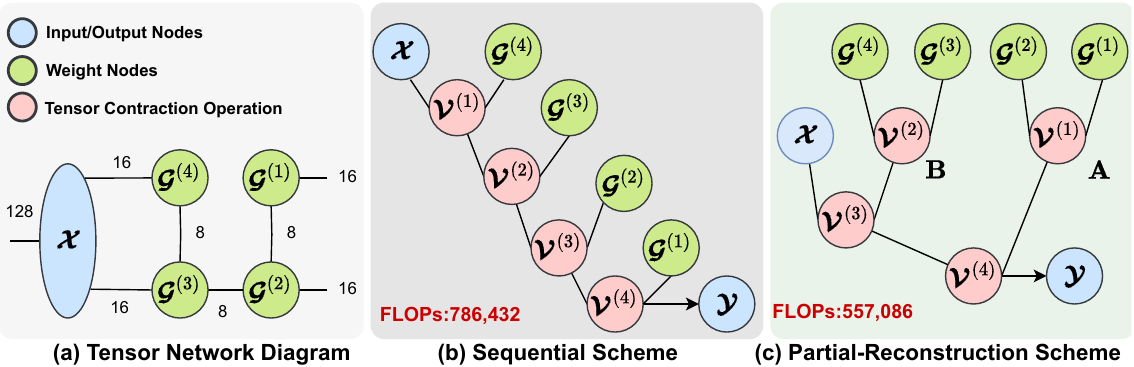}
  \caption{Example computing schemes for a TT-Layer.}
  \label{fig:tt_contraction}
    \vspace{-10pt}
\end{figure}
\subsection{Partial-Reconstruction Computing Scheme for TT Layers}

% By applying the TT decomposition, the standard linear layer $\mat{Y} = \mat{X} \mat{W}^T$ is replaced with a TT-Layer. Specifically, the weight matrix $\mat{W} \in \mathbb{R}^{M \times N}$ is tensorized into a series of TT-cores $\{\ten{G}^{(i)}\}_{i=1}^{2d}$, with corresponding TT-ranks $\{r_i\}_{i=0}^{2d}$, as defined in Eq.~\eqref{eq:tt}.
% The input and output are reshaped into high-dimensional tensors: $\ten{X} \in \mathbb{R}^{B \times n_1 \cdots n_d}$ and $\ten{Y} \in \mathbb{R}^{B \times m_1 \cdots m_d}$, where $B$ is the batch size, and ${n_k}$, ${m_k}$ are the mode sizes of the input and output tensors, respectively.
% The TT-Layer can be expressed as:
% \begin{equation}
%   \ten{Y}_{[b,i_1\cdots i_d]} = \sum_{j_1 ... j_d}  \ten{G}^{(1)}{[i_1]} \ten{G}^{(2)}{[i_2]} ... \ten{G}^{(2d)}{[j_d]} \ten{X}_{[b,j_1 ... j_d]},
% \end{equation}
% where $\ten{G}^{(i)}[i_k] \in \mathbb{R}^{r_{i-1} \times r_i}$ is the $i_k$-th slice of the TT-core $\ten{G}^{(i)}$ by fixing its second index as $i_k$.

Fig.~\ref{fig:tt_contraction} presents an example of a TT-Layer with $B = 128$, $m_i = [16, 16]$, $n_i = [16, 16]$, and $r_i = [1, 8, 8, 8, 1]$.
For clarity, Fig. \ref{fig:tt_contraction} (a) visualizes the TT-Layer computation as a tensor network.
Each node represents a multi-dimensional tensor, while the connecting edges denote contracted dimensions, often associated with tensor multiplication. The tensor network for a TT-Layer consists of $(2d + 1)$ nodes and requires $2d$ tensor contraction operations to compute the final output.

Given a TT-Layer, there are many feasible computing orders. With perfect numerical precisions, different computing schemes are equivalent in terms of the final result.
However, this equivalency is {\bf not guaranteed} when quantization is applied.
The placement of quantization operators within the TT-Layer critically affects the final output, as quantization errors accumulate along the contraction path.
% In general, the activations and internal tensors are more sensitive to the quantization than weight nodes.
For example, a typical computing scheme is to perform the tensor contraction in descending order of the core tensor indices, referred to as the \textbf{sequential} scheme in Fig. \ref{fig:tt_contraction} (b) \cite{gong2023ette}. This scheme  imposes a contraction depth of $2d$ on the activation path, leading to significant precision degradation.
%Another scheme is to reconstruct $\mat{W}$ from core tensors and then process it as a standard neural layer  (Fig.~\ref{fig:tt_contraction}(c)). This approach minimizes activation depth to 1, mitigating error accumulation. However, the reconstruction process incurs much higher computational complexity\cite{tensorly,t3f}.

To address the trade-off between accuracy and efficiency in quantized TT-Layers, we propose a novel computation strategy termed the Partial-Reconstruction Scheme (PRS).
As illustrated in Fig. \ref{fig:tt_contraction} (c), the computation involves three steps:
\begin{itemize}
  \item[\ding{202}] Output dimension reconstruction: core tensors associated with the output dimensions (i.e., $\ten{G}^{(1)}, \ten{G}^{(2)}, \cdots, \ten{G}^{(d)}$) are contracted to form a partial weight matrix $\mat{A} \in \mathbb{R}^{r_d \times M }$.
  \item[\ding{203}] Input dimension reconstruction: core tensors associated with the input dimensions (i.e., $\ten{G}^{(d+1)}, \ten{G}^{(d+2)}, \cdots, \ten{G}^{(2d)}$) are contracted to construct another partial matrix $\mat{B} \in \mathbb{R}^{N \times r_d }$.
  \item[\ding{204}] Input contraction: the input node is sequentially contracted with $\mat{A}$ and $\mat{B}$ to obtain the final output: $ \mat{Y} = \mat{X} \times\mat{B} \times \mat{A}$.
\end{itemize}
% Besides, the step \ding{202} and \ding{203} can be executed in parallel, offering additional opportunities for hardware-level acceleration.

\begin{figure}[tb]
  \centering
  \includegraphics[width=\columnwidth]{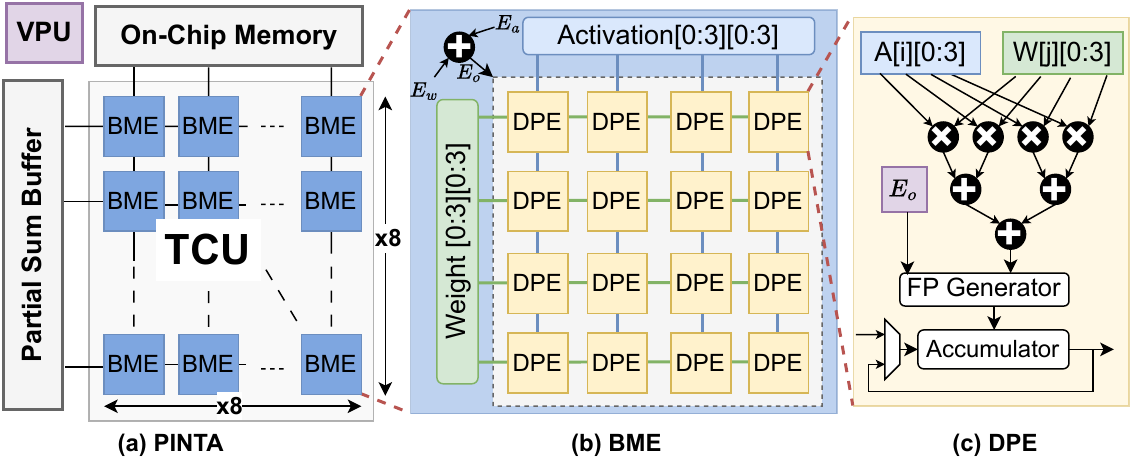}
  \caption{Overview of PINTA architecture.}
  \label{fig:arch}
    \vspace{-10pt}
\end{figure}

\section{Hardware Architecture}

To validate the performance of the proposed training framework, we designed and implemented an efficient hardware architecture, namely \textbf{\underline{P}}hysics-\textbf{\underline{I}}nformed \textbf{\underline{N}}eural \textbf{\underline{T}}raining \textbf{\underline{A}}ccelerator (PINTA).
As depicted in Fig.~\ref{fig:arch}, PINTA comprises a Tensor Contraction Unit (TCU), a 32-way vector processing unit (VPU), a partial sum buffer, and on-chip memory banks.

TCU executes variable-precision tensor contractions using an $8\times8$ array of Block Matrix computation Engines (BMEs). This array is organized as a transposable systolic array \cite{lu2025fetta} to support the flexible dataflows required by TT training.
Each BME performs a block of SMX matrix multiplication under a shared exponent, issuing $4\times4\times4 = 64$ MACs per cycle.
Internally, a BME is composed of a $4\times 4$ array of Dot-Product Engines (DPEs).

Each DPE performs four INT4$\times$INT4 multiplications per cycle along the reduction dimension within each data blocks.
The immediate result is converted to a floating-point format using a shared exponent ($E_o = E_w+E_a$) and is then accumulated with previous partial sums.
When operand precision is greater than 4 bits, the DPE accumulates partial sums bit-serially. This mechanism enables precision-scalable arithmetic while preserving block floating-point semantics. For instance, an INT8$\times$INT8 operation requires four cycles to complete. Upon completion of each tensor contraction, the VPU processes subsequent operations, such as nonlinear activations or quantization.

\section{Experiments}

\subsection{Evaluation Setting}
We evaluate our method on three representative PDE benchmarks of varying dimensionality: the 2D Poisson equation, the 20D Hamilton–Jacobi–Bellman (HJB) equation, and the 100D Heat equation. All models are trained on a single NVIDIA RTX A6000 Ada GPU using the Adam optimizer with a learning rate of $1e{-3}$ for 1,000 iterations. The number of samples for Stein's estimator is set to 512.
%In fully quantized training, 8-bit precision is used for both weights and activations, while 12-bit precision is employed for gradients.

\textbf{Poisson Equation:}
\begin{equation}
  \left\{
  \begin{aligned}
      \Delta u(x) &= g(x) && x \in \Omega \\
      u(x) &= h(x) && x \in \partial \Omega,
  \end{aligned}
  \right.
\end{equation}\label{eq:poisson}
where $\Omega = [0, 1] \times [0, 1]$ is the domain of the problem, $g(x) = -\text{sin}(x_1 + x_2)$ and $h(x) = 1/2 \text{sin}(x_1 + x_2)$.
The base network is a 4-layer MLP with 256 neurons and \textit{tanh} activation function.

\textbf{HJB Equation:}
\begin{equation}
  \left\{
  \begin{aligned}
    &\partial_t u(x,t) + \Delta u(x,t) - 0.5 \left\| \nabla_x u(x,t) \right\|_2^2 = -2, \\
    &u(x,1) = \|x\|_1, \quad x \in [0,1]^{20}, \quad t \in [0,1].
  \end{aligned}
  \right.
\end{equation}\label{eq:hjb}
A 4-layer MLP with 512 neurons is utilized \cite{zhao2025scalable}.

\textbf{Heat Equation:}
\begin{equation}
  \left\{
    \begin{aligned}
    u_t(x,t) &= \Delta u(x,t) && x \in B(0,1),\; t \in (0,1) \\
    u(x,0) &= \|x\|^2 / 2N && x \in B(0,1) \\
    u(x,t) &= t + 1 / 2N && x \in \partial B(0,1),\; t \in [0,1]
    \end{aligned}
    \right.
\end{equation}\label{eq:heat}
The base network is a 4-layer MLP with 256 neurons.
%and \textit{tanh} activation function in each layer.

\subsection{Accuracy Evaluation}

We compare our proposed method against two primary baselines: a full-precision, full-rank model trained with Automatic Differentiation (AD-FP-FR) and its counterpart trained with the Stein's Estimator (SE-FP-FR).
The accuracy results are summarized in Table \ref{tab:final_res}, where we report mean squared error (MSE), $\ell_1$ relative error, and $\ell_2$ relative error.
The results demonstrate that our tensorized, fully quantized training strategy achieves accuracy comparable to, or in some cases exceeding, the uncompressed baselines.

We also analyze the effect of the TT-rank selection. A smaller rank yields a higher model compression ratio but can also lead to a larger approximation error. This highlights the crucial trade-off between model efficiency and predictive accuracy that must be considered when selecting the rank.

% For the 2D Poisson  problem, our method achieves slightly higher error than the AD-Full and SE-Full baselines. While for the 20D HJB  and 100D Heat  problems, our method yields even better accuracy than baselines.

\setlength{\tabcolsep}{6pt}
\begin{table}[t]
  \centering
  \caption{Experimental results on PDE solving.}
  \label{tab:final_res}
\renewcommand{\arraystretch}{1.2}
\resizebox{0.99\columnwidth}{!}{%
\begin{tabular}{c|c|c|ccc}
  \toprule
                            & \multicolumn{2}{c|}{Method}   & MSE      & $\ell_1$ Rel. Error & $\ell_2$ Rel. Error \\ \hline
  \multirow{5}{*}{\textbf{Poisson}}    & \multicolumn{2}{c|}{AD-FP-FR}  & 1.15E-03 & 2.45E-03      & 2.86E-03      \\
                              & \multicolumn{2}{c|}{SE-FP-FR}  & 1.59E-03 & 3.10E-03      & 3.93E-03      \\ \cline{2-6}
                              & \multirow{3}{*}{Ours} & $R=8$  & 3.28E-03 & 6.83E-03      & 8.16E-03      \\
                              &                       & $R=16$ & 2.27E-03 & 4.42E-03      & 5.65E-03      \\
                              &                       & $R=32$ & 1.96E-03 & 3.75E-03      & 4.88E-03      \\ \hline
  \multirow{5}{*}{\textbf{HJB}}        & \multicolumn{2}{c|}{AD-Full}  & 9.11E-02 & 6.51E-03      & 8.61E-03      \\
                              & \multicolumn{2}{c|}{SE-Full}  & 4.18E-02 & 3.37E-03      & 3.95E-03      \\  \cline{2-6}
                              & \multirow{3}{*}{Ours} & $R=8$  & 4.65E-02 & 3.83E-03      & 4.38E-03      \\
                              &                       & $R=16$ & 1.93E-02 & 1.58E-03      & 1.82E-03      \\
                              &                       & $R=32$ & 2.47E-02 & 1.68E-03      & 2.33E-03      \\ \hline
  \multirow{5}{*}{\textbf{Heat}}       & \multicolumn{2}{c|}{AD-Full}  & 4.28E-03 & 7.33E-03      & 7.33E-03      \\
                              & \multicolumn{2}{c|}{SE-Full}  & 3.66E-03 & 6.28E-03      & 6.28E-03      \\  \cline{2-6}
                              & \multirow{3}{*}{Ours} & $R=8$  & 4.37E-03 & 7.46E-03      & 7.46E-03      \\
                              &                       & $R=16$ & 4.93E-03 & 8.48E-03      & 8.48E-03      \\
                              &                       & $R=32$ & 4.60E-03 & 7.86E-03      & 7.86E-03      \\ \bottomrule
  \end{tabular}%
  }
\end{table}

\subsubsection{Ablation Study}
% \zz{Better to show this subsection first.}
% We conduct break-out studies to demonstrate the effectiveness of each component in the proposed framework.
The $\ell_2$ relative error of different design choices, using a TT-rank of 16, are summarized in Table \ref{tab:ablation}.
The SE-NaiveQuant method, which directly quantizes activations, exhibits significantly higher error compared to other configurations.
In contrast, the SE-DiffQuant method successfully preserves the accuracy of the full-precision model.
%  by employing difference-based quantization.

The SE-TT-Seq-DiffQuant variant, which uses a sequential contraction order, suffers from substantial error accumulation due to the sequential contraction order in the TT-Layer, resulting in degraded performance. By adopting  PRS, the SE-TT-PRS-DiffQuant method  mitigates this issue and achieves a more favorable trade-off between accuracy and efficiency.

\begin{table}[t]
\centering
\caption{$\ell_2$ relative error of different methods.}
\label{tab:ablation}
\renewcommand{\arraystretch}{1.2}
\resizebox{\columnwidth}{!}{%
\begin{tabular}{cccc}
\toprule
Problem            & Possion 2D & HJB 20D  & HEAT 100D   \\ \midrule
\rowcolor[HTML]{EFEFEF}
AD-FP-FR            & 2.86E-03  & 8.61E-03 & 7.33E-03 \\
SE-FP-FR            & 3.93E-03  & 3.95E-03 & 6.28E-03 \\
\rowcolor[HTML]{FFA07A}
SE-NaiveQuant      & 3.19E-01  & 3.82E-02 & 5.90E-02 \\
SE-DiffQuant       & 2.21E-03  & 4.15E-03 & 6.45E-03 \\
SE-FP-R16              & 5.93E-03  & 5.68E-3 & 7.39E-03  \\
\rowcolor[HTML]{FFA07A}
SE-TT-Seq-DiffQuant & 1.23E-02 & 2.00E-02 & 1.78E-01 \\
\rowcolor[HTML]{99E2B4}
\makecell{SE-TT-PRS-DiffQuant} & 5.65E-03 & 1.82E-03 & 8.48E-03 \\
\bottomrule
\end{tabular}%
}
\vspace{-5pt}
\end{table}

\begin{figure}[tb]
  \centering
  \includegraphics[width=0.49\textwidth]{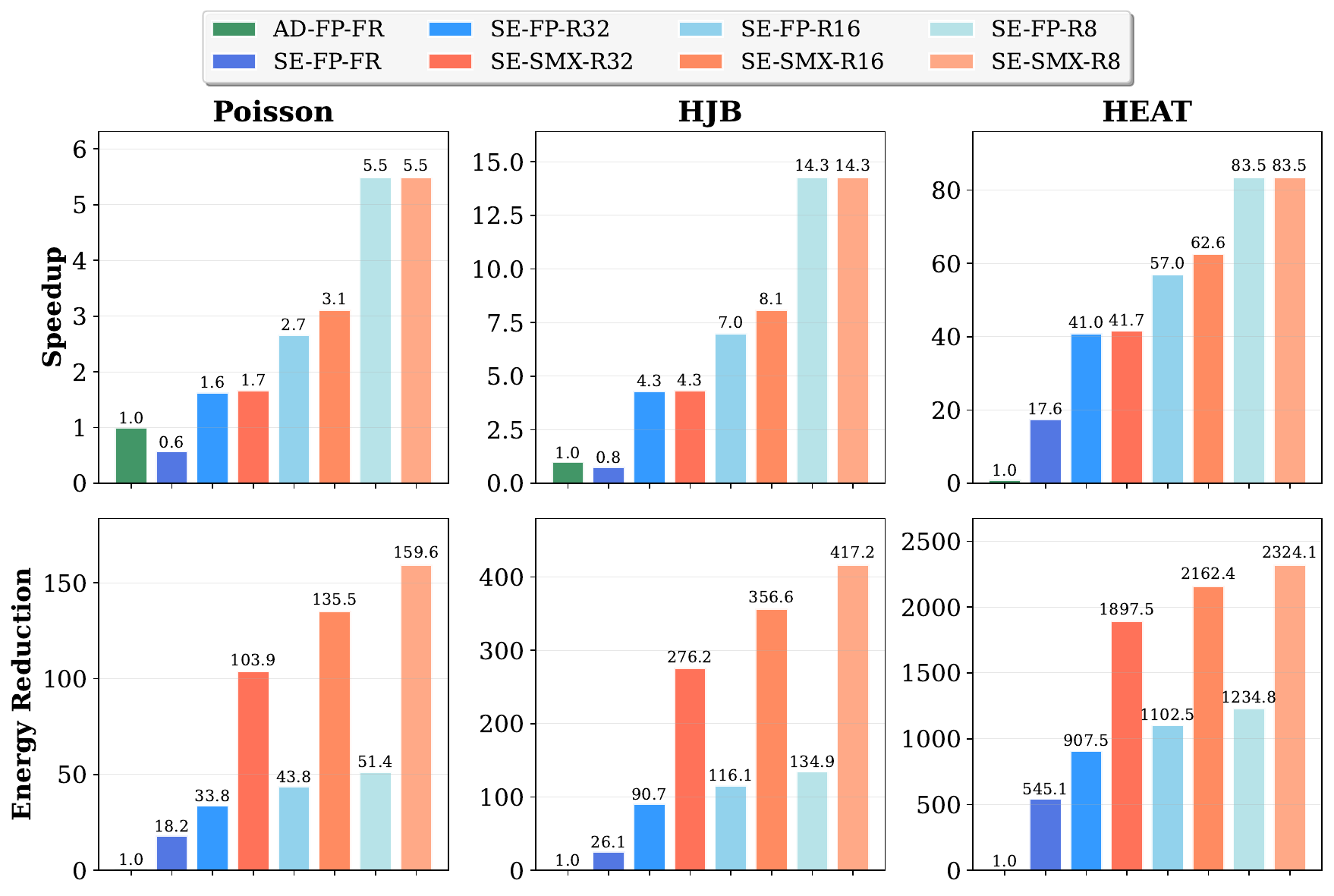}
  \caption{Hardware performance comparison under different settings.} %\zz{What does the y-axis mean?}}
  \label{fig:speedup}
    \vspace{-10pt}
\end{figure}
\vspace{-5pt}
\subsection{Hardware Performance}

 We implemented PINTA in SystemVerilog RTL and synthesized it using the ASAP 7nm technology.
 The resulting prototype occupies $0.442$ mm$^2$, contains 384-KB on-chip memory,  and operates at 1.0 GHz.
%  To evaluate performance and energy consumption, we developed a cycle-accurate simulator.
 The area and power of on-chip SRAM were modeled using PCACTI~\cite{shafaei2014fincacti}, while the HBM2 memory was modeled with an access energy of 3.9 pJ/bit and a bandwidth of 256 GB/s.

 The performance of AD-FP-FR baseline was evaluated on an NVIDIA RTX 4060 laptop GPU with 70W TDP.
 Training the model on the Poisson, HJB, and Heat equations required 25.96s, 134.30s, and 450.63s, respectively.
 To assess the full-precision SE models, We also implement a systolic array architecture with $32\times 32$ FP32 MACs units. This was used  benchmark SE models with different TT-ranks, including SE-FP-FR and SE-FP-R*.
 In contrast, models using our proposed framework (SE-SMX-R*) were evaluated on the PINTA accelerator.

Fig. \ref{fig:speedup} shows a comparison of training speed and energy consumption across the different configurations.
Lower TT-ranks (from R=32 down to R=8) and our SMX configuration (SE-SMX-R*) yield monotonic performance improvements across all benchmarks.
At a rank of $R=8$, the proposed system achieves  $5.5\times$ to $83.5\times$ speedups, together with $159.6\times$ to $2324.1\times$ reductions in energy consumption compared to AD-FP-FR baseline running on GPU.
Compared to the SE-FP-FR configurations, SE-SMX-R8 improves speed by $4.7\times$ to $18.3\times$ and energy efficiency  $4.3\times$ to $16.0\times$, respectively.
For any given rank, PINTA achieves $1.88\times$ to $3.10\times$ energy reductions over its full-precision SE counterpart (SE-FP-R*).

% The latencies for AD and SE baselines are measured on a GPU platform, while the proposed method is measured by the theoretical effective FLOPs. As shown in the figure, our method  achieves a complexity reduction of $1928 \times  \sim 32745 \times $ relative to AD-Full, and $ 90 \times \sim 361 \times $ relative to SE-Full. These results highlight the substantial efficiency gains of our approach, demonstrating its potential to significantly accelerate training for high-dimensional PDE problems without compromising accuracy.

\section{Conclusion}
This paper presents an efficient framework for solving high-dimensional PDEs with PINNs by integrating a Stein's derivative estimator, tensor-train decomposition, and fully quantized training.
Our framework utilizes a mixed-precision, square-block MX-INT format to ensure high representational fidelity and memory efficiency. We introduce two novel techniques to maintain accuracy in this low-precision setting: a difference-based quantization scheme that preserves the sensitivity of the Stein's estimator, and a partial-reconstruction scheme that mitigates error accumulation during TT-Layer computations. Experimental results demonstrate that our proposed framework significantly improves training efficiency over full-precision baselines without compromising predictive accuracy.

\newpage

%\newpage
% \zz{(1) remove underlines in the references. (2) pay attention to the upper cases in paper titles (e.g., FPGA --> \{FPGA\}) in the bibligraphy file.}
\bibliographystyle{ieeetr}
\bibliography{ref}
\end{document}